\documentclass[preprint,12pt]{elsarticle}

\usepackage{amssymb}
\usepackage{amsmath}
\usepackage{graphicx}%
\usepackage{float}%
\usepackage{placeins}%
\usepackage{multirow}%
\usepackage{amsmath,amssymb,amsfonts}%
\usepackage{amsthm}%
\usepackage{mathrsfs}%
\usepackage[title]{appendix}%
\usepackage{xcolor}%
\usepackage{textcomp}%
\usepackage{manyfoot}%
\usepackage{booktabs}%
\usepackage{algorithm}%
\usepackage{algorithmicx}%
\usepackage{algpseudocode}%
\usepackage{listings}%

\usepackage{subcaption}
\usepackage{changepage}
\usepackage{xcolor}
\usepackage{tikz}
\usepackage{lineno}

\journal{IEEE Transactions on Medical Imaging}

\begin{document}

\begin{frontmatter}

%% Title, authors and addresses

%% use the tnoteref command within \title for footnotes;
%% use the tnotetext command for theassociated footnote;
%% use the fnref command within \author or \affiliation for footnotes;
%% use the fntext command for theassociated footnote;
%% use the corref command within \author for corresponding author footnotes;
%% use the cortext command for theassociated footnote;
%% use the ead command for the email address,
%% and the form \ead[url] for the home page:
%% \title{Title\tnoteref{label1}}
%% \tnotetext[label1]{}
%% \author{Name\corref{cor1}\fnref{label2}}
%% \ead{email address}
%% \ead[url]{home page}
%% \fntext[label2]{}
%% \cortext[cor1]{}
%% \affiliation{organization={},
%%             addressline={},
%%             city={},
%%             postcode={},
%%             state={},
%%             country={}}
%% \fntext[label3]{}

\title{PRISM: a framework harnessing unsupervised visual representations and textual prompts for explainable MACE survival prediction from cardiac cine MRI}

%% use optional labels to link authors explicitly to addresses:
\author[label1,label2,label3]{Haoyang Su\fnref{equal}}
\fntext[equal]{These authors contributed equally to this work.}

\author[label4]{Jin-Yi Xiang\fnref{equal}}

\author[label2,label3,label4]{Shaohao Rui}

\author[label2,label5]{Yifan Gao}

\author[label4]{Xingyu Chen}

\author[label4]{Tingxuan Yin}

\author[label3,label4]{Shaoting Zhang}

\author[label2,label3]{Xiaosong Wang\corref{cor1}}
\ead{wangxiaosong@pjlab.org.cn}

\author[label4]{Lian-Ming Wu\corref{cor1}}
\ead{wlmssmu@126.com}

\cortext[cor1]{Corresponding authors}

\affiliation[label1]{organization={Fudan University},
            city={Shanghai},
            country={China}}

\affiliation[label2]{organization={Shanghai Innovation Institute},
            city={Shanghai},
            country={China}}

\affiliation[label3]{organization={Shanghai Artificial Intelligence Laboratory},
            city={Shanghai},
            country={China}}

\affiliation[label4]{organization={Shanghai Jiao Tong University},
            city={Shanghai},
            country={China}}

\affiliation[label5]{organization={University of Science and Technology of China},
            % addressline={School of Biomedical Engineering (Suzhou), Division of Life Science and Medicine},
            city={Hefei},
            country={China}}

%% Abstract
\begin{abstract}
%% Text of abstract
Accurate prediction of major adverse cardiac events (MACE) remains a central challenge in cardiovascular prognosis. We present PRISM (Prompt-guided Representation Integration for Survival Modeling), a self-supervised framework that integrates visual representations from non-contrast cardiac cine magnetic resonance imaging with structured electronic health records (EHRs) for survival analysis. PRISM extracts temporally synchronized imaging features through motion-aware multi-view distillation and modulates them using medically informed textual prompts to enable fine-grained risk prediction. Across four independent clinical cohorts, PRISM consistently surpasses classical survival prediction models and state-of-the-art (SOTA) deep learning baselines under internal and external validation. Further clinical findings demonstrate that the combined imaging and EHR representations derived from PRISM provide valuable insights into cardiac risk across diverse cohorts. Three distinct imaging signatures associated with elevated MACE risk are uncovered, including lateral wall dyssynchrony, inferior wall hypersensitivity, and anterior elevated focus during diastole. Prompt-guided attribution further identifies hypertension, diabetes, and smoking as dominant contributors among clinical and physiological EHR factors.
\end{abstract}

% %%Graphical abstract
% \begin{graphicalabstract}
% %\includegraphics{grabs}
% \end{graphicalabstract}

% %%Research highlights
% \begin{highlights}
% \item We propose a motion-aware multi-view distillation framework for unsupervised representation learning from 4D cardiac cine MRI sequences, which extracts spatiotemporal features through teacher-student knowledge transfer between long-axis and short-axis views, enabling effective capture of myocardial dynamics.

% \item We introduce an EHR-attention directive mechanism that enables cross-modal alignment between imaging and clinical data through medically informed prompts, coupled with a BiPromptSurv strategy that identifies cohort-specific EHR feature groups for improved survival prediction across diverse clinical centers.

% \item Our framework addresses a critical clinical need by enabling accurate prediction of MACE, a crucial composite endpoint in cardiovascular disease encompassing mortality, myocardial infarction, revascularization, and heart failure, from non-contrast cardiac cine MRI without requiring mask-level annotations, providing a practical solution for cardiovascular risk stratification essential for clinical decision-making.
% \end{highlights}

%% Keywords

\end{frontmatter}

%% Add \usepackage{lineno} before \begin{document} and uncomment 
%% following line to enable line numbers
%% \linenumbers

%% main text
%%

\section{Introduction}
Major adverse cardiovascular events (MACE), encompassing outcomes such as cardiovascular death, myocardial infarction, and hospitalization for heart failure, remain a significant cause of morbidity and mortality globally~\cite{Visseren2021ESC, Martin2024Heart, Reyes2023MACE, Xu2024Attention}. As a composite endpoint, accurate prediction of MACE is essential for informing clinical decisions, optimizing patient outcomes, and capturing clinically significant deterioration, which correlates with long-term prognosis. This makes MACE prediction central to cardiovascular risk stratification~\cite{Bosco2021MajorAC,RAZIPOUR2025AI}. Consequently, many predictive approaches, such as AI-driven methods~\cite{Schrempf2021,pi2025moscardcausalreasoning,rui2025cardiocothierarchicalreasoningmultimodal,hu2024aipredictioncardiovascularevents,song2024pericoronaryadiposetissuefeature,STAREcho2023,Tariq2024, Xu2024Attention, Kim2024SelfAttentionMACE, Zhang2024MLMACEPCI, liu2025dynamicsurvivalpredictionusing}, have become increasingly critical in MACE prediction in recent years.

\begin{figure}[h!]
    \centering
    \includegraphics[width=1\linewidth]{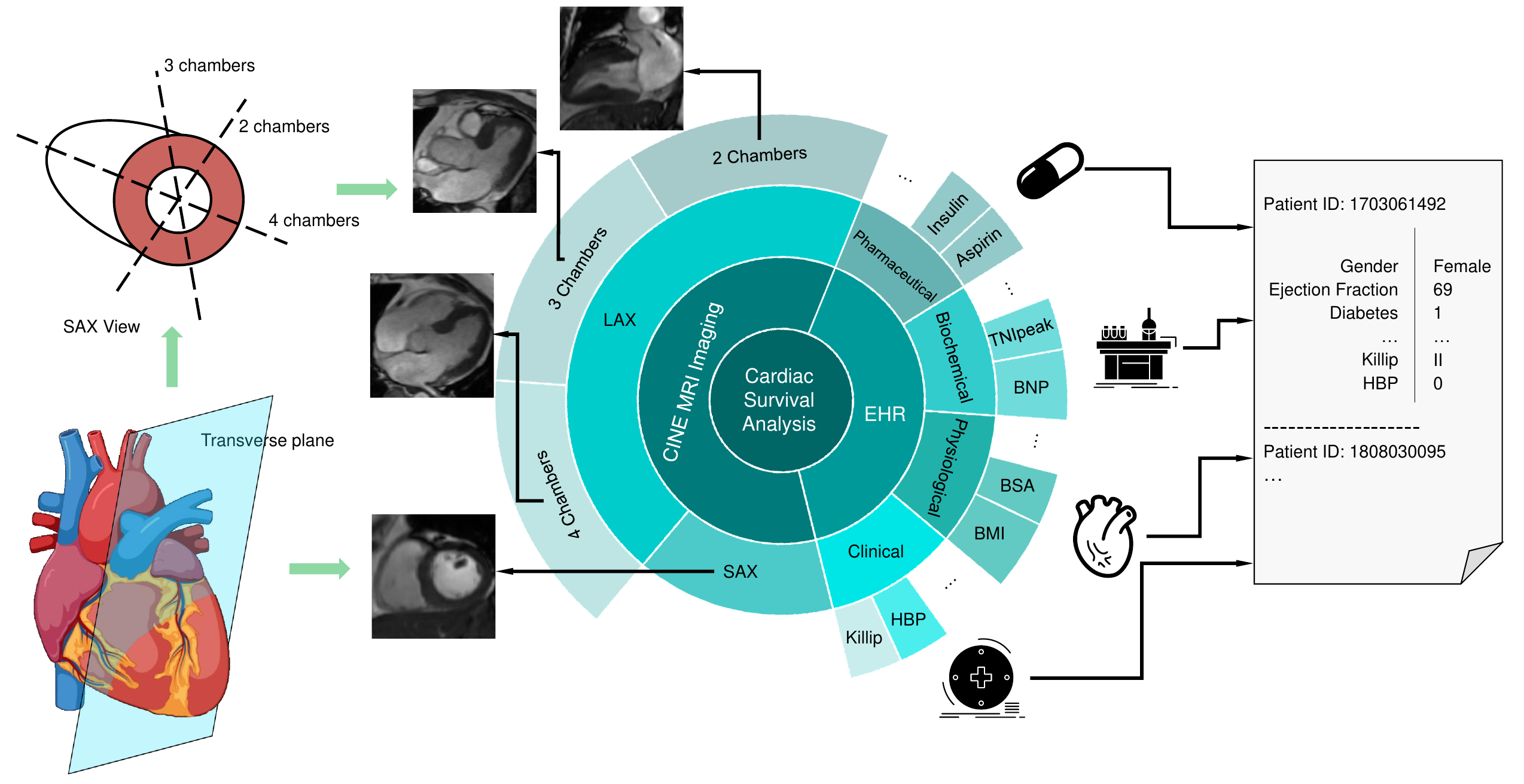}
    \caption{Overview of the cardiac survival analysis dataset featuring non-contrast cine MRI in short and long axis views and patient-level EHRs subcategorized into pharmaceutical, biochemical, physiological, and  clinical domains.}
    \label{dataset_overview}
\end{figure}

Traditional approaches~\cite{Nagpal2021DSM, Meier2020DeepSurvivalGastric, Cox2018CoxPH} often purely rely on manually curated features derived from electronic health records (EHRs), which, despite their accessibility, offer refined but limited physiological resolution and may fail to reflect the dynamic nature of cardiac function. Alternatively, methods based on histopathological images~\cite{terada2022histopathological,udin2024differential, Meier2020DeepSurvivalGastric,Wang2024Gptomic, Chen2023Transformer}, while providing high-resolution cellular information, are inherently static and dimensionally restricted, making them less suitable for capturing organ-level or temporal characteristics relevant to adverse cardiovascular outcomes. These limitations hinder the extraction of comprehensive prognostic representations. In contrast, cardiac cine magnetic resonance imaging (MRI) provides organ-scale, high-resolution, and temporally dynamic information~\cite{PETITJEAN2011169, KLEM2012408,JACOB2025101967,fu2025versatilefoundationmodelcine}, enabling the discovery of more expressive biomarkers associated with MACE~\cite{Li2025CineRadiomicsMACE, Konst2023CMRMINOCA, kato2021prognostic}. In this context, as Fig.~\ref{dataset_overview} shows, we exploit the complementary strengths of cine MRI and structured EHR data to construct a unified multimodal dataset tailored for integrative prognostic modeling.

Nevertheless, significant challenges remain. Contrast-enhanced acquisitions are often unavailable, and most medical images lack mask-level annotations, limiting the feasibility of expert-supervised modeling in real-world clinical settings. As a result, unsupervised paradigms become a prevailing approach in high-dimensional imaging representation learning~\cite{Xu2025Generalizable3DSSL, zhou2023selfpretrainingmaskedautoencoders, zhou2022preservationallearningimprovesselfsupervised,Williams2023Unsupervised, su2025ctslcodebookbasedtemporalspatiallearning}. However, unlike easily obtainable and superficial information contained in EHRs, conventional unsupervised image-to-image frameworks are confined to learning image-level representations, which, due to their inability to correlate with key attributes in EHRs, hinder the discovery of more granular biological biomarkers.

\begin{figure}[h!]
\begin{subfigure}[b]{1\textwidth}
  \centering
  \includegraphics[width=\textwidth]{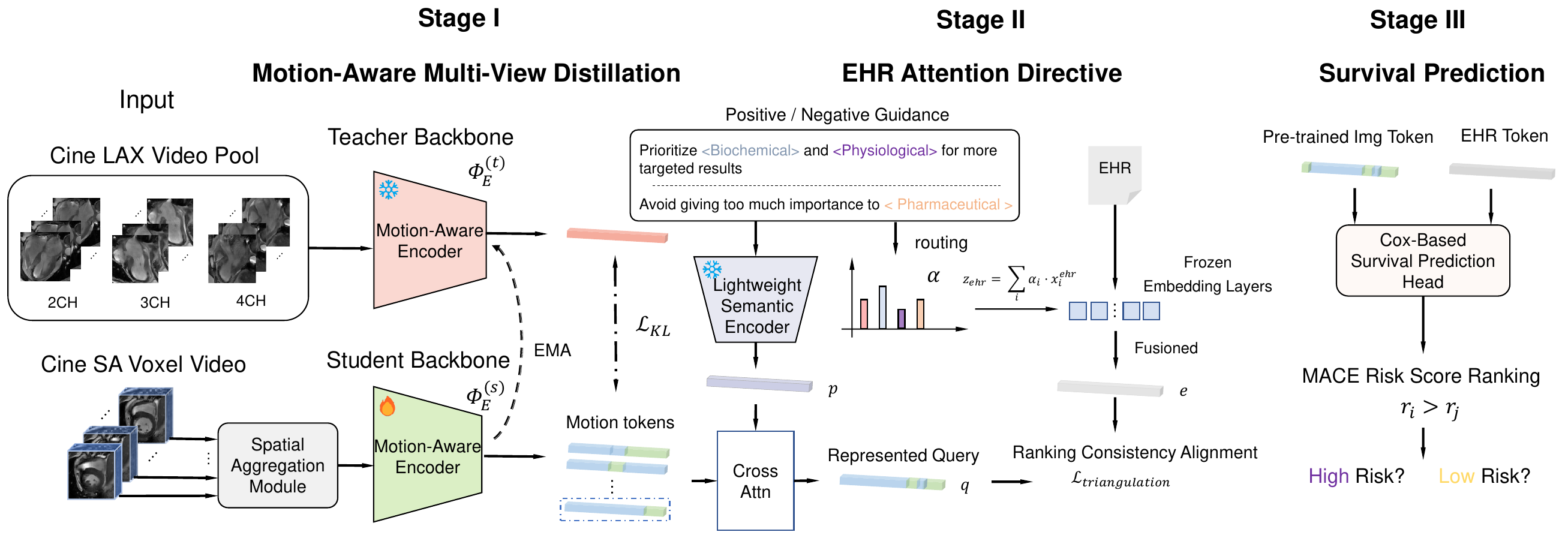}
  \caption{}
\end{subfigure}
\begin{subfigure}[b]{1\textwidth}
  \centering
  \includegraphics[width=\textwidth]{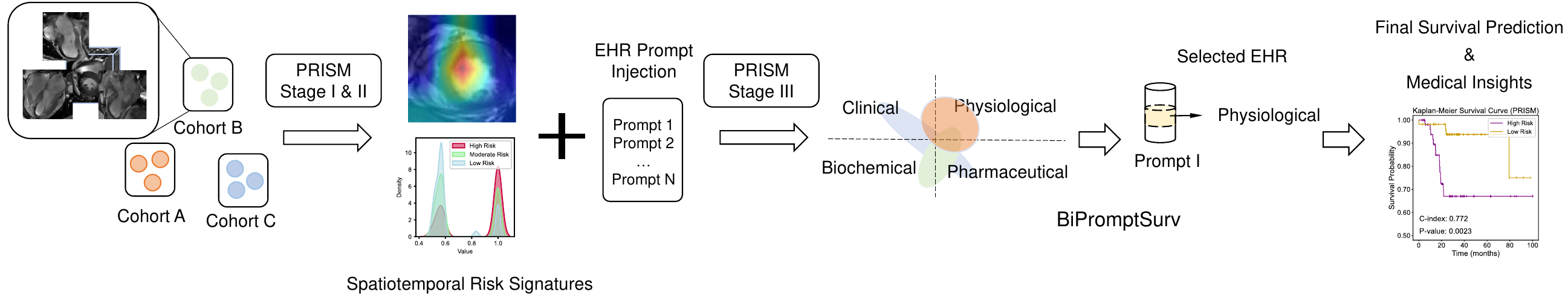}
  \caption{}
\end{subfigure}
    \caption{Design of the proposed framework and pipeline. (a), PRISM's three-stage survival analysis framework. Multi-view cine MRI images passed through a teacher-student distillation network to obtain representation tokens enriched with spatiotemporal features, aligned hierarchically with EHR features under the influence of medically informed prompt aggregation. The refined image features are ultimately aggregated with EHR features in Stage III to yield survival analysis results. (b), Medical insights discovery pipeline based on PRISM. Deriving the association between spatiotemporal patterns of ventricular myocardium and MACE risk from the learned heatmap distributions, and exploring EHR features aiding in MACE assessment based on the BiPromptSurv strategy.}
    \label{framework_pipeline}
\end{figure}

To address these limitations, we propose PRISM, a prompt-guided representation integration framework for survival modeling, which jointly learns spatiotemporal representations from cine MRIs and enables their interaction with structured EHR data, as illustrated in Fig.~\ref{framework_pipeline}a. Image features are first extracted from cardiac cine sequences integrated with a multi-view distillation module, followed by alignment with EHR embeddings through a triangulation loss applied to the outputs of a student network. To further bridge the gap between image-derived features and downstream survival outcomes such as MACE, clinically meaningful prompts are incorporated to guide the model toward medically relevant semantics, thereby facilitating the discovery of prognostic imaging biomarkers indicative of cardiac risk. Building upon this, we derive clinically meaningful insights through the representations learned by PRISM by following the pipeline in Fig~\ref{framework_pipeline}b. The image-derived features are utilized to reflect differential survival risk and MACE incidence. Building on these features, we introduce BiPromptSurv in Stage III as an outcome-guided analytical strategy that integrates semantically grouped EHR categories as anchors. By aligning model attention with domain-informed subgroups, this approach enables the identification of prognostically informative signatures and offers interpretable medical insights.

\section{Materials and methods}
\subsection{Image preprocessing: ventricular motion focus}
To rigorously quantify myocardial dynamics across consecutive cine frames, we employed the Farneb\"ack optical flow algorithm~\cite{Farneb2003Two} on full cardiac volumes $\mathcal{V}^{(total)}$. This approach computes dense displacement fields that provide a continuous and high-resolution representation of local tissue motion, thereby capturing the spatiotemporal behavior of ventricular wall kinematics. The resulting displacement vector field was formulated as
\begin{equation}
\mathbf{F}^{(t)} = \Psi_{\mathrm{FB}}\left(\mathcal{I}_{t}, \mathcal{I}_{t+1}\right),
\label{equ:farneback}
\end{equation}
where $\mathbf{F}^{(t)}$ denotes the motion vector field between successive frames $\mathcal{I}_{t}$ and $\mathcal{I}_{t+1}$, and $\Psi_{\mathrm{FB}}$ represents the Farneb\"ack optical flow operator. This formulation allows for voxel-level characterization of inter-frame myocardial motion, which is subsequently refined by focusing on a standardized region-of-interest centered on the ventricular chambers with a standardized 96-pixel window encompassing the cardiac chambers, thereby focusing the analysis on ventricular myocardial motion while excluding signals from extracardiac structures.

\subsection{Stage I: motion-aware multi-view distillation}
\begin{figure}[h!]
    \centering
    \includegraphics[width=1\linewidth]{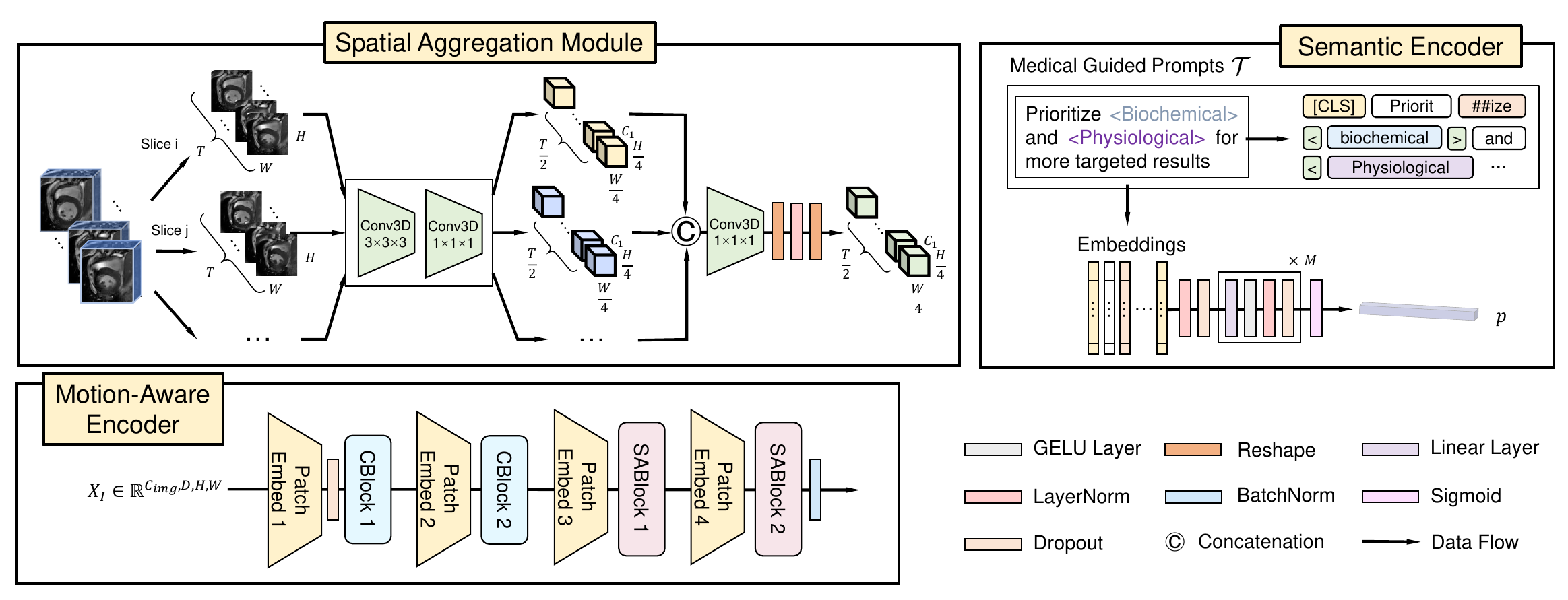}
    \caption{Details of model architecture and sub-modules. The Spatial Aggregation Module aggregates latent representations obtained from layer-wise convolutions over cardiac cine MRI sequences.
The Semantic Encoder encodes medical guided prompts $\mathcal{T}$ based on pretrained embeddings.
The Motion-Aware Encoder incorporates the CBlock and SABlock modules from the UniFormer~\cite{li2022uniformer} backbone.}
    \label{fig:arch_submodule}
\end{figure}

To achieve dimensional alignment with LAX, the SAX images are processed in a slice-wise manner by the Spatial Aggregation Module as Fig.~\ref{fig:arch_submodule} shows, yielding latent representations.
To capture comprehensive myocardial motion patterns, the framework implements a teacher-student architecture inspired by DINOv2~\cite{oquab2024dinov2learningrobustvisual}, where the backbone is inherited from UniFormer~\cite{li2022uniformer} for temporal modeling, as illustrated in Fig.~\ref{fig:arch_submodule}. The teacher pathway processes long-axis views comprising 2-chamber, 3-chamber, and 4-chamber orientations, generating multi-view feature embeddings that encapsulate global myocardial dynamics. The student pathway encodes spatially compressed short-axis sequences to extract condensed motion representations. Knowledge transfer is enforced by the distillation loss,
\begin{equation}
\mathcal{L}_{\mathrm{Stage_I}} = \tau^{2} D_{\text{KL}}\left( p^{(s)} \| \bar{p}^{(t)} \right) + \lambda \mathcal{L}_{\text{contrastive}},
\label{equ:stageI}
\end{equation}
where $\bar{p}^{(t)}$ represents the teacher's multi-view feature embeddings, $p^{(s)}$ denotes the student's short-axis representations, and $D_{\text{KL}}$ is the Kullback-Leibler divergence~\cite{Kullback1951OnIA}. The complementary contrastive term $\mathcal{L}_{\text{contrastive}}$, implemented via InfoNCE~\cite{chen2020simple}, enforces intra-patient phase-specific feature consistency while separating motion characteristics across different subjects. This formulation yields spatiotemporally coherent motion tokens that capture prognostic signatures of ventricular dysfunction and facilitate downstream modeling.

\subsection{Stage II: EHR-attention directive}
As Fig.~\ref{fig:arch_submodule} shows, to enable effective cross-modal alignment between clinical and imaging data, the model employs a routing module $\mathcal{R}$ that maps a medical prompt $\mathcal{T} \in \mathbb{V}^L$, where $\mathbb{V}$ denotes the clinical vocabulary and $L$ is the sequence length, into a sparse group attention vector,
\begin{equation}
\boldsymbol{\alpha} =  \mathcal{R}(\text{Embed}(\mathcal{T})),
\label{equ: alpha_embed}
\end{equation}
where each $\alpha_i$ is approximately 0, 0.5, or 1.

The architecture then branches into three parallel streams to establish cross-modal alignment. The Cross-Attention Fusion stream constructs query projections $\mathbf{Q} = \mathbf{P}\mathbf{W}_Q$ from BERT-encoded prompt embeddings~\cite{Koroteev2021BERT} and linearly projects spatial-temporal visual features $\mathbf{Z}^{(s)}$ to obtain keys and values $\mathbf{K} = \mathbf{V} = \mathbf{Z}^{(s)}\mathbf{W}_K$. The aligned image representation is computed as
\begin{equation}
\mathbf{Z}_{align} = \text{LayerNorm}\left( \mathbf{W}_p \cdot \mathbb{E}_l\left[ \mathbf{Q}_l \oplus \text{Attn}(\mathbf{Q}_l,\mathbf{K},\mathbf{V}) \right] \right),
\label{equ: cross-attn}
\end{equation}
where $\oplus$ denotes concatenation and $\mathbb{E}_l[\cdot]$ represents average pooling over prompt tokens. For EHR feature processing, each categorical variable $e_f$ in group $g$ is embedded via dedicated layers $\phi_f(\cdot)$ and aggregated with routing weights $\alpha_g$,
\begin{equation}
\mathbf{Z}_{ehr}^g = \alpha_g \cdot \mathbb{E}_{f \in S_g}\left[ \phi_f(e_f) \right], \quad \mathbf{Z}_{ehr} = \sum_{g=1}^G \mathbf{Z}_{ehr}^g,
\label{equ: routing_weight}
\end{equation}
where $S_g$ indexes the clinical feature subgroup. Finally, the Visual Anchor stream preserves the original spatial-temporal feature topology via
\begin{equation}
\mathbf{Z}_{ref} = \text{LayerNorm}\left( \mathbf{W}_r \cdot \mathbb{E}[\mathbf{Z}^{(s)}] \right),
\label{equ: ref}
\end{equation}
ensuring structural integrity while supporting subsequent alignment objectives.

\subsection{Alignment Objectives}
The dual-objective loss enforces both semantic alignment and structural preservation. The EHR-Consistent Triangulation Loss operates on valid triplets $(i,j,k)$ satisfying $\text{sim}_{EHR}(\mathbf{e}^i,\mathbf{e}^j) > \text{sim}_{EHR}(\mathbf{e}^i,\mathbf{e}^k) + \delta$ as defined in Equ.~\ref{equ: loss_tri},
\begin{equation}
\mathcal{L}_{tri} = \sum_{\mathcal{T}} \max\left( 0, \|\mathbf{Z}_{align}^i - \mathbf{Z}_{ehr}^j\|_2^2 - \|\mathbf{Z}_{align}^i - \mathbf{Z}_{ehr}^k\|_2^2 + \delta \right),
\label{equ: loss_tri}
\end{equation}
with margin $\delta=0.2$. The Topology Preservation Loss combining geometric constraints is calculated in Equ.~\ref{equ: pres},

\begin{equation}
\mathcal{L}_{pres} = \frac{1}{B}\sum_{i=1}^B  \|\mathbf{Z}_{align}^i - \mathbf{Z}_{ref}^i\|_2^2,
\label{equ: pres}
\end{equation}
where $B$ denotes the batch size. Then we obtain the composite objective in Equ.~\ref{equ: loss_tri_loss_pres},
\begin{equation}
\mathcal{L}_{total} = \mathcal{L}_{tri} + \beta \mathcal{L}_{pres}.
\label{equ: loss_tri_loss_pres}
\end{equation}

\subsection{Stage III: survival prediction}
To predict patient survival, clinical indicators $\mathbf{e} \in \mathbb{R}^{d_m}$ and pre-trained image tokens $\mathbf{r} = \mathbb{E}(Q_{img}) \in \mathbb{R}^{d_c}$ are concatenated into a fused representation,
\begin{equation}
\mathbf{x}_{\mathrm{fused}} = \mathrm{concat}[\mathbf{e}, \mathbf{r}] \in \mathbb{R}^{d_m + d_c},
\label{equ: fused_rep}
\end{equation}
integrating information from both clinical and imaging domains. A Cox proportional hazards (CoxPH) head is applied, where the coefficients $\beta_k$ automatically weight the contribution of each fused feature. The hazard function is expressed as
\begin{equation}
h(t | \mathbf{x}_{\mathrm{fused}}) = h_0(t) \exp \left( \sum_k \beta_k x_{\mathrm{fused}_k} \right),
\label{equ: Cox_h}
\end{equation}
with $h_0(t)$ representing the baseline hazard function. Training minimizes the negative log-partial likelihood,
\begin{equation}
\mathcal{L}_{\mathrm{Cox}} = - \sum_{i:\delta_i = 1} \left[ \mathbf{\theta}^\top \mathbf{x}_{\mathrm{fused}}^{(i)} - \log \sum_{j \in R(t_i)} \exp \left( \mathbf{\theta}^\top \mathbf{x}_{\mathrm{fused}}^{(j)} \right) \right] + \lambda \| \mathbf{\theta} \|_2^2,
\label{equ: Cox_loss}
\end{equation}
where $\mathbf{\theta} = [\beta_1, \ldots, \beta_m]^\top$ denotes the regression coefficients, $\delta_i$ indicates whether the event occurred for patient $i$, and $R(t_i)$ corresponds to the risk set at survival time $t_i$. This provides a principled framework to integrate clinical and imaging data for prognostic assessment.

\subsection{Evaluation metrics}
In this study, we utilized C-index~\cite{Harrell2001RegressionMS} and area under the curve (AUC) as primary evaluation metrics. The C-index assesses the discriminatory power of the model by measuring the proportion of all comparable subject pairs whose predicted risks are correctly ordered relative to their observed survival times. Formally, it is defined in Equ.~\ref{equ: Cindex},
\begin{equation}
\text{C-index} = \frac{\sum\limits_{i,j} \mathbb{I}(t_i < t_j) \mathbb{I}(r_i > r_j) \delta_i}{\sum\limits_{i,j} \mathbb{I}(t_i < t_j) \delta_i},
\label{equ: Cindex}
\end{equation}
where $\delta_i$ indicates the occurrence of the event of interest, and $r_i = \mathbf{\beta}^\mathrm{T} \mathrm{x}_{\mathrm{fused}}^{i}$ represents the risk score derived from the prognostic model. Complementarily, we incorporated AUC to quantify the model's sensitivity and specificity in stratifying patients with respect to event status across varying decision thresholds. The time-dependent AUC at time $t$ is defined as
\begin{equation}
\text{AUC}(t) = P(r_i > r_j | y_i(t) = 1, y_j(t) = 0),
\label{equ: AUC}
\end{equation}
where $y_i(t)$ denotes the event status of subject $i$ at time $t$, and $r_i$ represents the predicted risk score. This metric measures the probability that a subject experiencing an event at time $t$ has a higher risk score than a subject who remains event-free at that time point.

\subsection{Implementation details}
The model was implemented in the PyTorch environment and trained using NVIDIA-H100 GPUs with CUDA 12.2. Specifically, Stages I and III were conducted on a single H100 GPU, whereas Stage II utilized a 4-GPU H100 setup. The GPU-to-CPU ratio was maintained at 1 to 15. Optimization was performed using the Adam optimizer~\cite{Kingma2014AdamAM} with an initial learning rate of \(5 \times 10^{-5}\) and weight decay set to \(1 \times 10^{-5}\). A batch size of 16 was used throughout, and models were trained for 50 epochs with a StepLR learning rate scheduler.  The 4D cardiac cine MRI inputs were uniformly resized to dimensions of \(24 \times 24 \times 96 \times 96\), indicating depth, temporal frames, height, and width. A total of 41 EHR features were incorporated, with 579, 672, 406, and 313 samples included from GLCCM, AZCCM, RJCCM, and TJCCM, respectively.

Standard data partitioning schemes were employed, with a training-to-validation ratio of 7:3 for self-supervised learning during Stages I and II, and a split of 6:2:2 for training, validation, and testing sets in the survival prediction stage. In Stage II, each image sample was paired with 50 unique medical prompt variants. During survival prediction stage, grid search combined with stratified sampling was used for hyperparameter tuning, and feature selection was performed using stepwise regression and Lasso regularization within the CoxPH. The internal–external cross-validation (IECV)~\cite{steyerberg2016prediction} setting is implemented by fixing the same test set within each cohort. For internal validation, training and testing are performed on the individual cohort separately. For external validation, the model is trained on the combined training sets of the remaining cohorts and tested on the same test set as used in the internal validation.

\section{Results}
\subsection{Generalizability of PRISM across internal–external validation settings in multi-center cohorts}
To rigorously evaluate the robustness and generalizability of our proposed framework, we conducted a comprehensive IECV analysis across four independent cohorts. PRISM was comparatively assessed against classical statistical models, neural network–based approaches, and contemporary SOTA survival methodologies using the concordance index (C-index) and time-dependent AUC as primary performance metrics. The comparator models encompassed the CoxPH~\cite{Cox2018CoxPH}, established as the clinical gold standard for survival analysis, as well as DeepSurv~\cite{Meier2020DeepSurvivalGastric} and Deep Survival Machines (DSM)~\cite{Nagpal2021DSM}, designed to capture non-linear risk functions from tabular EHR data. Represented SOTA models in 3D medical imaging–based survival analysis, including Sparse BagNet and SurvRNC, were utilized to benchmark the performance of PRISM.  Furthermore, to evaluate cross-paradigm efficacy within unsupervised representation learning, PRISM was benchmarked against two prominent self-supervised frameworks, namely PCRL~\cite{zhou2022preservationallearningimprovesselfsupervised} and 3DINO~\cite{Xu2025Generalizable3DSSL}, which pretrained on large-scale, multimodal medical imaging datasets.

\begin{table}[h!]
  \centering
  \caption{Performance comparison (C-index) of different models across cohorts under IECV settings.}
  \resizebox{\textwidth}{!}{%
    \begin{tabular}{lcccc}
    \toprule
    Model & GLCCM & AZCCM & RJCCM  & TJCCM  \\
    \midrule
          & \multicolumn{4}{c}{\textbf{Internal}} \\
    \midrule
    CoxPH & \underline{0.698}$\pm$0.071 & \underline{0.760}$\pm$0.020 & 0.568$\pm$0.040 & 0.619$\pm$0.091 \\
    DSM   & 0.572$\pm$0.079 & 0.656$\pm$0.048 & 0.536$\pm$0.076 & 0.532$\pm$0.037 \\
    DeepSurv & 0.553$\pm$0.095 & 0.616$\pm$0.092 & 0.545$\pm$0.104 & 0.593$\pm$0.087 \\
    SurvRNC & 0.641$\pm$0.091 & 0.716$\pm$0.063 & 0.583$\pm$0.109 & 0.579$\pm$0.043 \\
    Sparse BagNet & 0.536$\pm$0.050 & 0.547$\pm$0.110 & 0.532$\pm$0.146 & 0.495$\pm$0.102 \\
    3DINO & 0.573$\pm$0.047 & 0.588$\pm$0.049 & \underline{0.614}$\pm$0.055 & 0.634$\pm$0.115 \\
    PCRL  & \textbf{0.701}$\pm$0.052 & 0.749$\pm$0.044 & 0.526$\pm$0.089 & \underline{0.649}$\pm$0.098 \\
    PRISM (Ours) & 0.690$\pm$0.055 & \textbf{0.772}$\pm$0.035 & \textbf{0.617}$\pm$0.033 & \textbf{0.775}$\pm$0.037 \\
    \midrule
          & \multicolumn{4}{c}{\textbf{External}} \\
    \midrule
    CoxPH & 0.708$\pm$0.055 & 0.671$\pm$0.076 & 0.599$\pm$0.039 & 0.499$\pm$0.081 \\
    DSM   & 0.575$\pm$0.038 & 0.621$\pm$0.054 & 0.463$\pm$0.101 & 0.472$\pm$0.111 \\
    DeepSurv & 0.559$\pm$0.125 & 0.579$\pm$0.171 & 0.553$\pm$0.139 & 0.565$\pm$0.160 \\
    SurvRNC & 0.686$\pm$0.079 & \textbf{0.717}$\pm$0.044 & 0.582$\pm$0.038 & 0.513$\pm$0.093 \\
    Sparse BagNet & 0.563$\pm$0.077 & 0.459$\pm$0.040 & 0.492$\pm$0.077 & 0.559$\pm$0.103 \\
    3DINO & 0.587$\pm$0.030 & 0.566$\pm$0.050 & 0.581$\pm$0.080 & \underline{0.581}$\pm$0.120 \\
    PCRL  & \textbf{0.714}$\pm$0.049 & 0.609$\pm$0.046 & \underline{0.605}$\pm$0.053 & 0.573$\pm$0.021 \\
    PRISM (Ours) & \underline{0.710}$\pm$0.010 & \underline{0.712}$\pm$0.031 & \textbf{0.609}$\pm$0.077 & \textbf{0.701}$\pm$0.075 \\
    \bottomrule
    \end{tabular}%
    }
  \label{tab: main_results1} %
\end{table}%

\begin{table}[h!]
  \centering
  \caption{Performance comparison (AUC) of different models across cohorts under IECV settings.}
  \resizebox{\textwidth}{!}{%
    \begin{tabular}{lcccc}
    \toprule
    Model & GLCCM & AZCCM & RJCCM & TJCCM \\
    \midrule
          & \multicolumn{4}{c}{\textbf{Internal}} \\
    \midrule
    CoxPH & \underline{0.693}$\pm$0.083 & \underline{0.709}$\pm$0.042 & 0.574$\pm$0.049 & 0.601$\pm$0.074 \\
    DSM   & 0.562$\pm$0.060 & 0.626$\pm$0.058 & 0.520$\pm$0.065 & 0.521$\pm$0.038 \\
    DeepSurv & 0.543$\pm$0.105 & 0.583$\pm$0.070 & 0.526$\pm$0.112 & 0.574$\pm$0.095 \\
    SurvRNC & 0.645$\pm$0.075 & 0.682$\pm$0.077 & 0.583$\pm$0.121 & 0.552$\pm$0.061 \\
    Sparse BagNet & 0.507$\pm$0.054 & 0.562$\pm$0.084 & 0.542$\pm$0.130 & 0.509$\pm$0.140 \\
    3DINO & 0.586$\pm$0.049 & 0.558$\pm$0.039 & \textbf{0.620}$\pm$0.052 & 0.600$\pm$0.119 \\
    PCRL  & \textbf{0.699}$\pm$0.075 & 0.699$\pm$0.051 & 0.500$\pm$0.087 & \underline{0.657}$\pm$0.092 \\
    PRISM (Ours) & 0.689$\pm$0.036 & \textbf{0.714}$\pm$0.016 & \underline{0.593}$\pm$0.030 & \textbf{0.760}$\pm$0.066 \\
    \midrule
          & \multicolumn{4}{c}{\textbf{External}} \\
    \midrule
    CoxPH & 0.669$\pm$0.072 & \underline{0.676}$\pm$0.067 & \textbf{0.609}$\pm$0.033 & 0.474$\pm$0.076 \\
    DSM   & 0.569$\pm$0.026 & 0.606$\pm$0.055 & 0.479$\pm$0.079 & 0.463$\pm$0.114 \\
    DeepSurv & 0.552$\pm$0.108 & 0.599$\pm$0.130 & 0.551$\pm$0.118 & 0.559$\pm$0.132 \\
    SurvRNC & 0.683$\pm$0.073 & \textbf{0.678}$\pm$0.050 & 0.571$\pm$0.059 & 0.481$\pm$0.095 \\
    Sparse BagNet & 0.552$\pm$0.087 & 0.426$\pm$0.016 & 0.505$\pm$0.070 & 0.559$\pm$0.135 \\
    3DINO & 0.595$\pm$0.012 & 0.547$\pm$0.043 & 0.574$\pm$0.075 & \underline{0.600}$\pm$0.145 \\
    PCRL  & \textbf{0.713}$\pm$0.048 & 0.582$\pm$0.042 & \underline{0.594}$\pm$0.051 & 0.544$\pm$0.035 \\
    PRISM (Ours) & \underline{0.692}$\pm$0.019 & \underline{0.676}$\pm$0.028 & 0.589$\pm$0.091 & \textbf{0.657}$\pm$0.130 \\
    \bottomrule
    \end{tabular}%
    }
  \label{tab: main_results2} %
\end{table}%

Table~\ref{tab: main_results1} and Table~\ref{tab: main_results2} present the detailed comparison results, with the best and second-best performances for each cohort are highlighted in bold and underlined, respectively. PRISM consistently demonstrated superior performance in survival analysis across both in-domain and out-of-domain validations. Compared to models rooted in clinical tabular modeling paradigms such as CoxPH, DeepSurv, and DSM, PRISM demonstrated marked gains particularly in scenarios characterized by multimodal feature heterogeneity, highlighting that image-driven alignment of latent EHR prognostic factors underpins its enhanced effectiveness. While traditional models generally exhibited sharp performance decay in out-domain settings in the RJCCM and TJCCM cohorts, PRISM retained stable performance under distribution shift. Furthermore, when benchmarked against advanced SSL frameworks pretrained on large-scale medical imaging corpora, including PCRL and 3DINO, PRISM consistently exhibited higher discriminative capacity across both low- and high-risk strata. To further evaluate the significance of these results across random seeds, we combined the \textit{p}-values obtained from each seed using Fisher's method~\cite{Edwards2005Landmark}. The test statistic was computed as \(X^2 = -2 \sum_{i=1}^{k} \ln(p_i)\), which follows a \(\chi^2\) distribution with \(2k\) degrees of freedom. Detailed \textit{p}-values for each comparison are provided in Supplementary Table~B.1 and Supplementary Table~B.2.

\begin{figure}[h!]
  \centering
  \includegraphics[width=\textwidth]{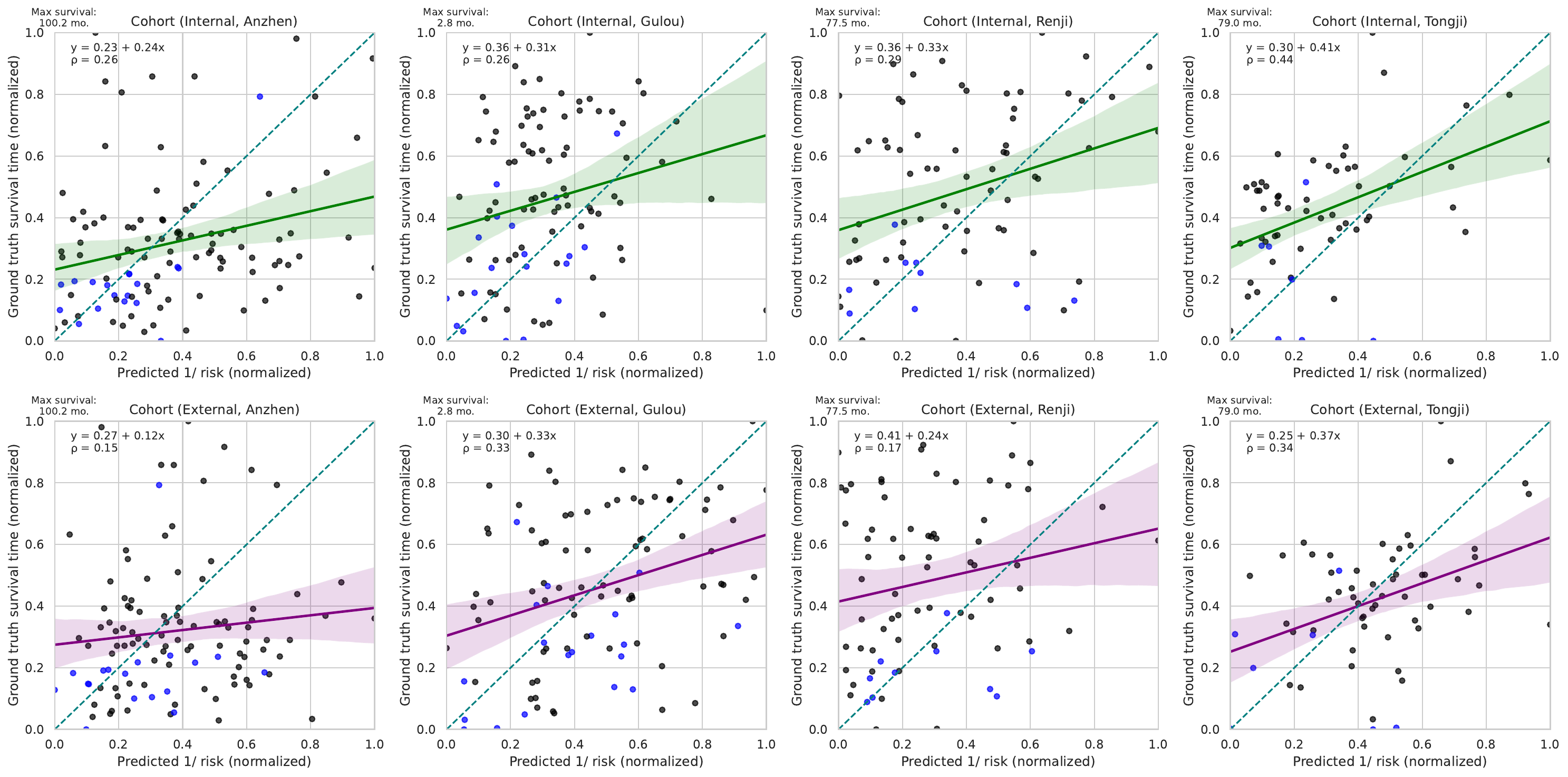}
  \caption{\textbf{Model survival analysis performance under the IECV setting.} Regression analysis in the interval-external cross-validation setting, with model-predicted inverse risk (normalized) on the horizontal axis versus ground-truth survival time (normalized) on the vertical axis. Top panels represent internal validation and bottom panels show external validation in correlated cohorts. Cases with MACE=0 are denoted in black, while those with MACE=1 are indicated in blue.}
  \label{fig:boxplots_regression}
\end{figure}

As illustrated in Fig.~\ref{fig:boxplots_regression}, the linear relationship between log-transformed inverse risk scores from our prognostic model and observed survival times in right-censored specimens further confirms the discriminative efficacy of computed risk in characterizing survival distributions. Specimens with MACE=1 consistently occupy high-risk positions across test sets in IECV setting. Critically, validation under external-center training conditions manifested only a minor slope reduction in average relative to internal validation. Specifically, the confidence band in the external GLCCM cohort exhibits progressive convergence with extended survival times, yielding statistically narrower 95\% confidence intervals versus the green internal band. This demonstrates time-invariant precision in survival estimation and temporal stability within the IECV experimental setting.

\begin{figure}[h!]
    \centering
    \includegraphics[width=1\linewidth]{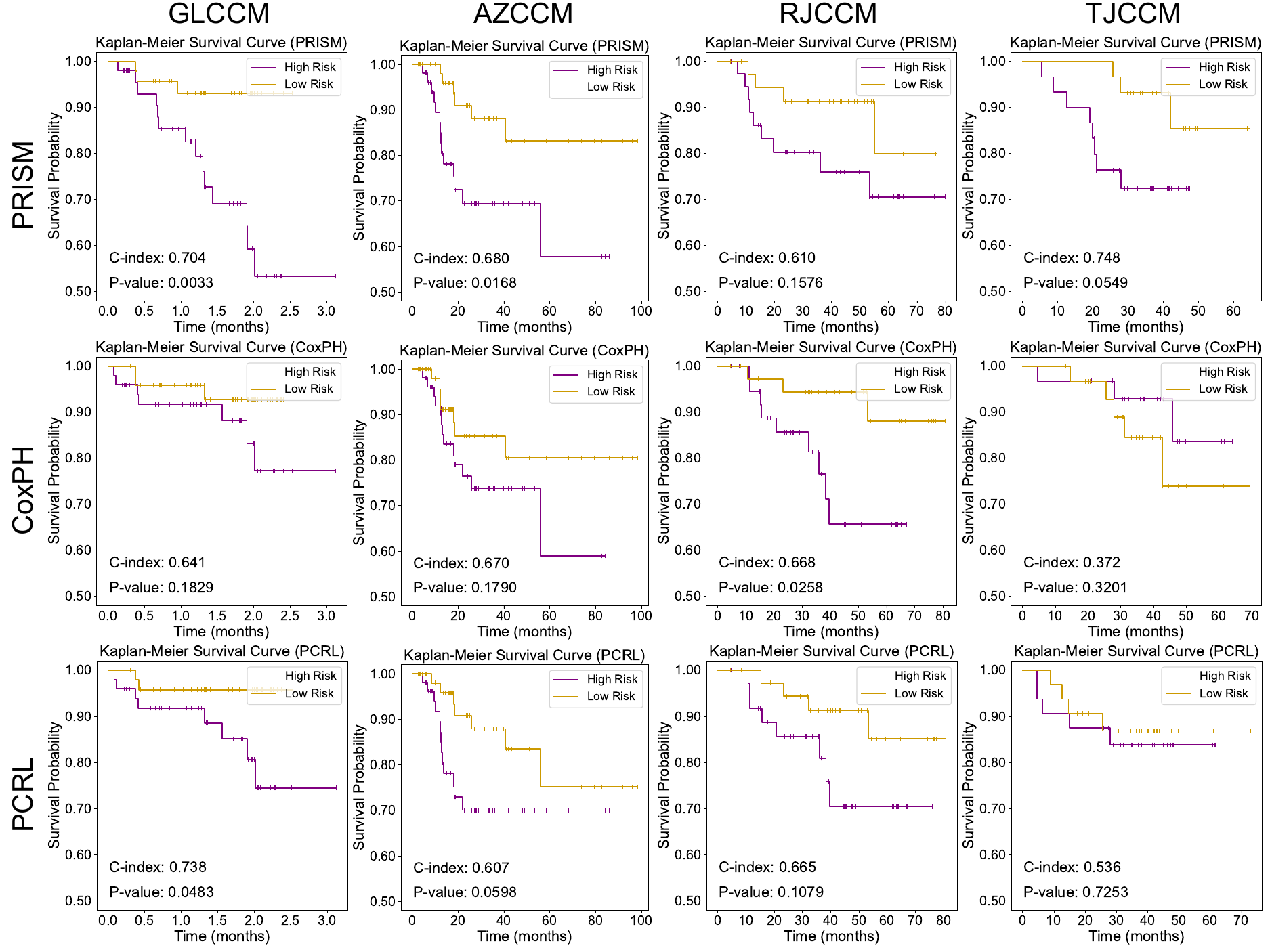}
    \caption{Kaplan–Meier survival plots across four cohorts (GLCCM, AZCCM, RJCCM, TJCCM) and three modeling strategies (PRISM, CoxPH, PCRL). \textit{p}-value is incorporated as a metric to assess the statistical significance and reliability.}
    \label{fig:KM}
\end{figure}

As illustrated in Fig.~\ref{fig:KM}, Kaplan–Meier survival curves~\cite{Kaplan1992} are presented for our proposed models, the clinical gold-standard CoxPH, and the strongest-performing comparative model PCRL under out-of-domain evaluation settings. The purple and yellow curves represent the high- and low-risk populations, respectively, stratified by the median of the model-predicted risk scores.

Examining the horizontal axis, due to the recency of data collection and limited follow-up windows, the GLCCM cohort exhibits relatively short survival durations within three months. This compressed time span leads to more confident \textit{p}-values across models. As the survival time extends in other cohorts, the uncertainty in MACE risk prediction increases, as reflected by an increase in the \textit{p}-value. This is also reflected by the flattening or overlapping of the high-risk survival curves in certain circumstances, such as PCRL on TJCCM. In contrast, PRISM consistently yields the smallest \textit{p}-values across cohorts, while the average \textit{p}-value of PRISM is approximately 16.19 times smaller than that of PCRL under the same target test set, indicating a much higher level of confidence to performance degradation over extended follow-up periods.

\subsection{Visual signatures of MACE in cardiac motility}

\begin{figure}[htbp!]
    \centering
    \begin{subfigure}[b]{1\textwidth}
        \centering
        \includegraphics[width=\textwidth]{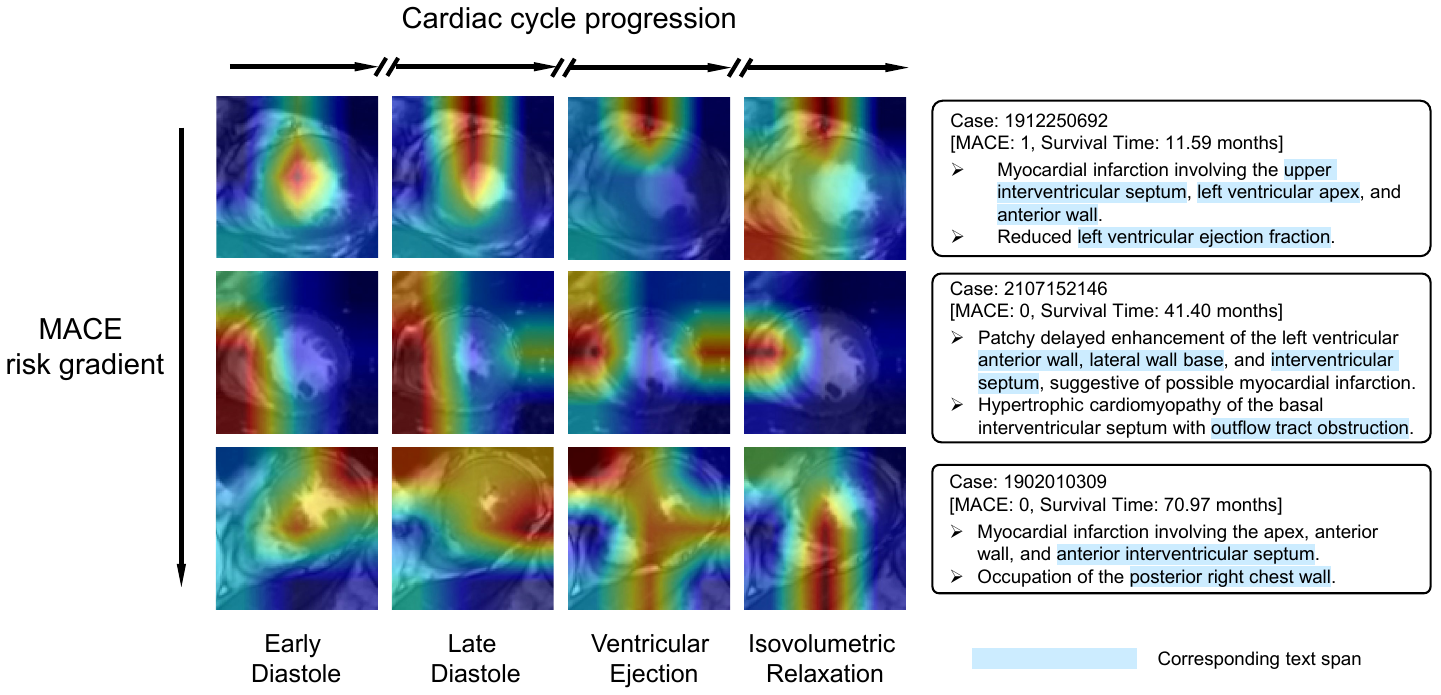}
        \caption{}
    \end{subfigure}
    \begin{subfigure}[b]{0.9\textwidth}
        \centering
        \includegraphics[width=\textwidth]{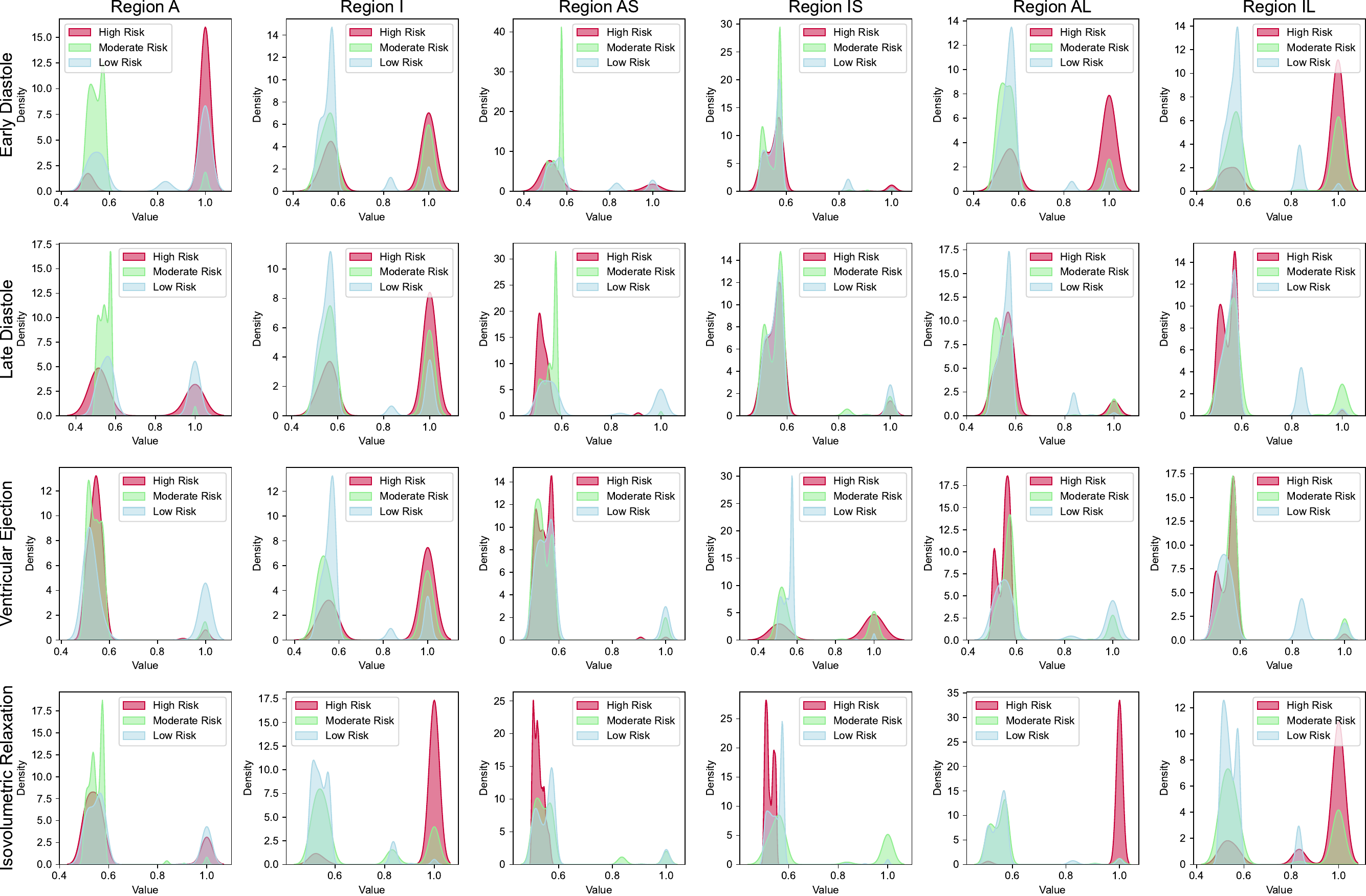}
        \caption{}
    \end{subfigure}
    \caption{\textbf{Model attention and heatmap analysis across cardiac cycle phases and MACE risk.} (a), Attention rollout of image tokens under stage II with horizontal axis representing typical cardiac cycle phases (early diastole, late diastole, ventricular ejection, isovolumetric relaxation) and vertical axis indicating increasing MACE risk levels. Text spans highlighted in light blue showing strong concordance between model attention and radiological diagnosis. (b), Spatiotemporally separated kernel density estimates of model-predicted heatmap distributions across three risk levels, six myocardial segments (A, I, AS, IS, AL, IL), and four cardiac cycle periods (early diastole, late diastole, ventricular ejection, isovolumetric relaxation), in the SAX view.}
    \label{fig: Cardiac_cycle_heatmap-KDE_plot}
\end{figure}

As Fig.~\ref{fig: Cardiac_cycle_heatmap-KDE_plot}a shows, with attention rollout applied during early diastole, late diastole, ventricular ejection, and isovolumetric relaxation, cross-attention distribution between the image-derived features and a medically guided prompt directing analytical capacity toward physiological aspects was visualized, revealing distinct phase-specific patterns of myocardial focus that corresponded to radiologically recognized regions of diagnostic relevance.

Radiological diagnostic reports established by board-certified physicians were served as reference standards for attention rollout validation. These expert-annotated reports were rigorously excluded from model training data throughout the unsupervised learning phase. For the subfigures, the spatial correspondence between cardiac anatomy and highlighted attention regions shows broad consistency with the described canonical structures, such as the four myocardial walls.

While the right ventricle itself was not directly implicated in the reported infarction, the attention pattern with extensive coverage over the right ventricular region in the upper subfigure may reflect secondary spatial proximity or structural interplay between the right heart border and adjacent thoracic pathology. This observation suggests that the model may possess the capacity to flag clinically relevant extracardiac abnormalities through contextual feature attribution.

On the other hand, representing a high-risk case with MACE and significantly reduced survival, the lower subfigure demonstrates progressive spatial redistribution of model attention. Following diastole, attention migrates from the left ventricle toward the anterior wall, subsequently returning to the left ventricular region post-systole and ejection, with eventual propagation toward the interventricular septum. This spatiotemporal trajectory aligns with diagnostic evidence of anterior wall myocardial infarction and reflects the model's sensitivity to dynamically coupled biomarkers such as ejection fraction variations.

Spatiotemporal characterization of myocardial attention patterns in Fig.~\ref{fig: Cardiac_cycle_heatmap-KDE_plot}b reveals distinct risk-dependent signatures across cardiac phases and ventricular segments. The standard mid-ventricular segmentation scheme~\cite{Standardized2002American} was applied, dividing the left ventricular myocardium into six circumferential wall segments. Kernel density estimation~\cite{Murray1956Remarks} of normalized heatmap values demonstrates consistently localized attentional foci, with transitional intervals exhibiting markedly sparse distributions. This indicates pronounced gradient steepness rather than diffuse activation, reflecting the model's capacity for targeted anatomical discrimination. In lateral segments (AL, IL), high-risk specimens exhibit significant phase-dependent redistribution. Diastolic phases display intensified focus relative to systolic attenuation, suggesting impaired relaxation-contraction coupling as a pathophysiological marker. Conversely, the inferior wall (segment I) manifests persistent hyper-attention across all phases in high-risk specimens, establishing this territory as a critical discriminator of adverse outcomes.

Septal regions (AS, IS) demonstrate functional homogeneity, with unimodal distributions centered near neutral values irrespective of risk stratification. This absence of differential engagement indicates limited pathological specificity in septal mechanics. The anterior wall (segment A) exhibits divergent risk-specific signatures, with moderate-risk specimens concentrating in low-attention zones while high-risk counterparts show early diastolic predominance.

Collectively, these signatures delineate three cardinal risk indicators, including lateral dyssynchrony, inferior hypersensitivity, and anterior diastolic elevated focus. The phase-dependent redistribution of attentional focus within lateral segments among high-risk subjects corresponds to established evidence positioning lateral strain delay as a superior marker of pathological left ventricular dyssynchrony~\cite{kuznetsova2013association}. Concordant with our findings, anterior diastolic elevated focus substantiates prior observations of the anterior wall's pronounced vulnerability to diastolic impairment, particularly within metabolically compromised substrates, as documented in type 2 diabetes cardiomyopathy models~\cite{daniels2023myocardial}. Similarly, the inferior hypersensitivity identified in our computational framework recapitulates the predilection for inferior wall involvement in ischemic pathology, a territorially critical determinant of cardiac prognosis~\cite{warner2023inferior}, thereby underscoring its pathognomonic role in myocardial ischemia-infarction continua.

\subsection{BiPromptSurv-driven EHR analysis}
Intuitively, comprehensive EHR integration yields the most accurate MA-\allowbreak CE risk stratification. However, our BiPromptSurv framework empirically identifies cohort-preferred EHR feature groups, enabling prompt-guided secondary survival analysis. By isolating salient feature subsets, this approach improves risk stratification consistency under controlled conditions, underscoring the prognostic value of selective feature integration. We introduced a set of syntactically diverse prompts (Supplementary Fig.~A.1) absent from the model's pre-training corpus. 
The prompts were synthesized via DeepSeek-R1 70B~\cite{deepseekai2025deepseekr1incentivizingreasoningcapability}, generating a prompt distribution,
which extrapolated novel yet semantically aligned instructions from a small set of handcrafted exemplars. A lightweight routing module assigns these prompts to categorized EHR class weights. The routing module's performance in recognizing different medical semantics of EHR categories is presented in the UpSet plot in Supplementary Fig.~A.2.

\begin{figure}[h!]
    \centering
    \includegraphics[width=1\linewidth]{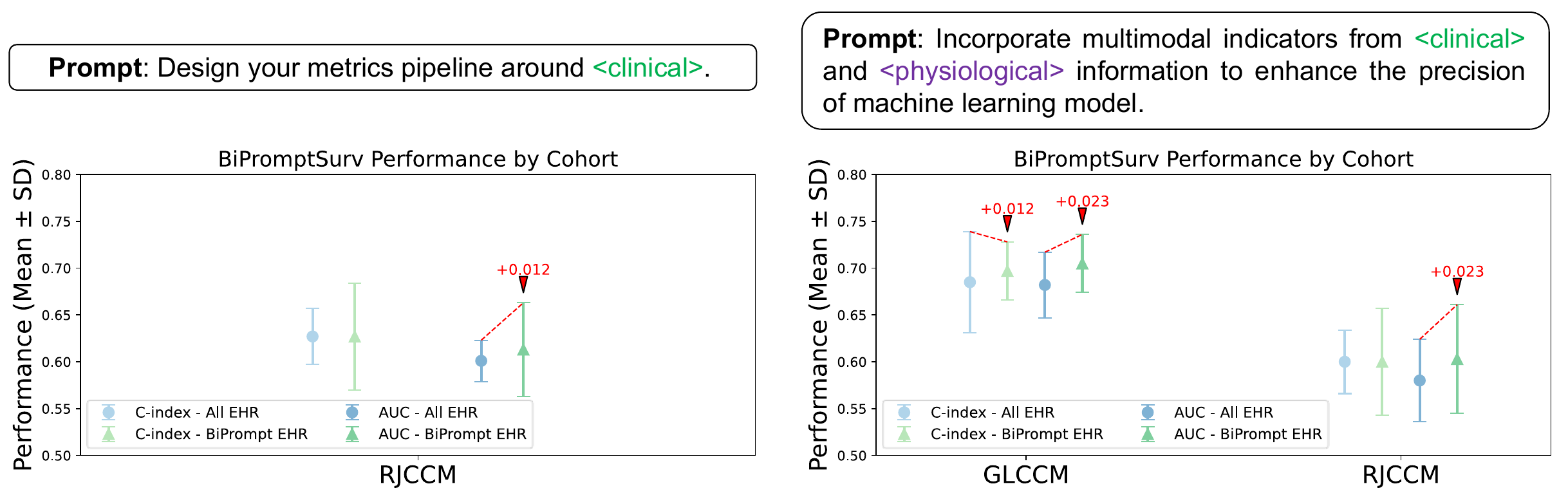}
    \caption{Examples of prompts under the BiPromptSurv pipeline along with result comparisons. Dot plots with error bars in light blue and light green across five seeds are displayed. Red arrows indicate the increase in mean corresponding metrics compared to the all EHR settings.}
    \label{fig:bipromptsurv_result}
\end{figure}

 Within the BiPromptSurv framework, a subset of generated prompts consistently improved risk stratification fidelity when applied to secondary survival analysis.
 Notably, as illustrated in Fig.~\ref{fig:bipromptsurv_result}, prompting with statements that focus on a single EHR category, such as \texttt{Design your metrics pipeline around <clinical>}, leads to a 1.2\% increase in the mean AUC for the RJCCM cohort, albeit at the cost of higher cross-seed variance.
Similarly, applying prompt like \texttt{Incorporate multimodal indicators from\linebreak   <clinical> and <physiological> information to enhance the prec-\linebreak ision of the machine learning model} led to improved survival prediction in both GLCCM and RJCCM cohorts by leveraging clinical and physiological features, outperforming models trained on the full feature set.

\begin{figure}[h!]
    \centering
    \begin{subfigure}[b]{0.9\textwidth}
        \centering
        \includegraphics[width=\textwidth]{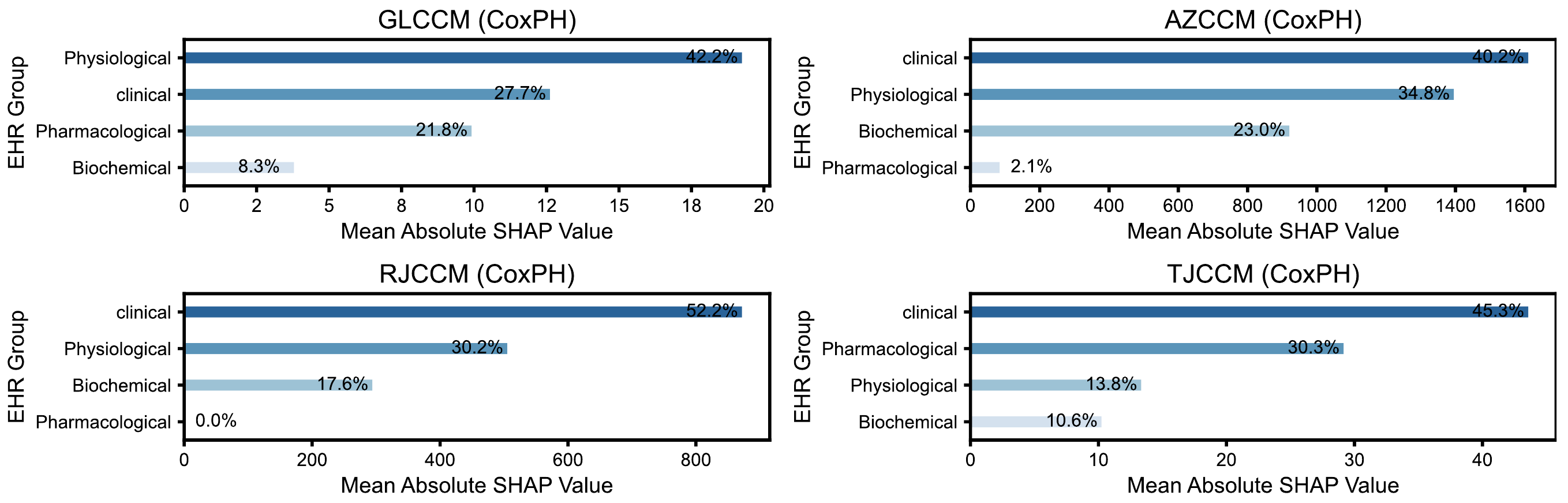}
        \caption{}
    \end{subfigure}
    \begin{subfigure}[b]{0.9\textwidth}
        \centering
        \includegraphics[width=\textwidth]{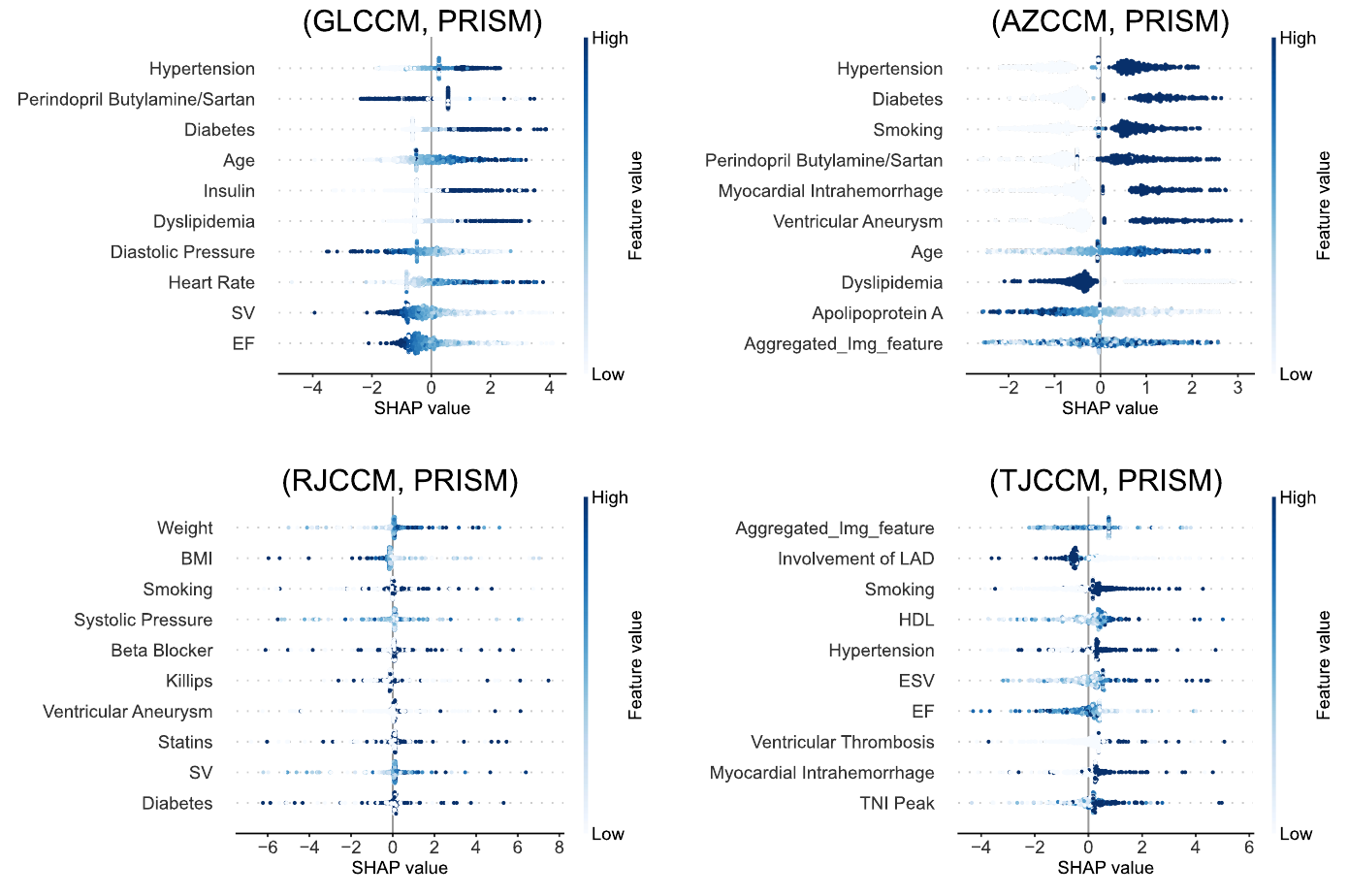}
        \caption{}
    \end{subfigure}
    \caption{Statistical analysis results based on PRISM Stage II and Stage III.
(a), Cluster attribution analysis based on CoxPH, showing contributions of each EHR category reflecting their survival analysis impact within the cohort.
(b), Fine-grained SHAP attribution analysis of PRISM. Top-10 ranked features are presented along with top-5 image features aggregated as \texttt{Aggregated\_Img\_feature}.}
\label{fig: BiPromptSurv}
\end{figure}

Building on this foundation, CoxPH model was employed, which serves as the gold standard and as the survival head within our framework, to conduct cohort-specific attribution analysis on the selected feature groups. Aggregated SHAP values were derived to demonstrate the contributions of different types of EHR features. As illustrated in Fig.~\ref{fig: BiPromptSurv}a, CoxPH attribution reveals that clinical and physiological features dominate the EHR contribution in RJCCM and GLCCM cohorts, accounting for approximately 82.4\% and 69.9\% of the total SHAP values, respectively. This concordance confirms BiPromptSurv's targeted feature selection efficacy while underscoring the interpretability of prompt-guided imaging features.

Furthermore, SHAP analysis was employed for fine-grained attribution in Fig.~\ref{fig: BiPromptSurv}b, aiming to uncover how our model's Stage II and III behaviors vary across different prompts and the respective contributions of specific categorized EHR features to survival analysis (the full EHR-level SHAP analysis is provided in Supplementary Fig.~A.3). Using PRISM, we identified and presented the top-10 most important features at the cohort level, among which is an \texttt{Aggregated\_Img\_feature} aggregated from the five core imaging features. Concordance was observed in GLCCM, where physiological indicators such as heart rate (HR), stroke volume (SV), and ejection fraction (EF), which are closely related to myocardial function, were included. This is consistent with the role of PRISM in extracting imaging features during Stage I. Meanwhile, clinical features such as hypertension, diabetes, and dyslipidemia also remained dominant. The same trend was observed in RJCCM, where physiological and clinical features accounted for 80\% of the top-10.
Additionally, the role of aggregated image features as the top-ranked primary contributors in the TJCCM cohort is prominently demonstrated in Fig.~\ref{fig:boxplots_regression}, further validating the effectiveness of the PRISM model in image analysis by achieving significant improvements over classical EHR models, SOTA methods, and unsupervised comparative approaches.

\subsection{Ablation Study}
Table~\ref{tab:ablation_c-index} and Table~\ref{tab:ablation_auc} present the ablation study results comparing PRISM with and without Stage II across all cohorts under IECV settings.
PRISM without Stage I represents an ablation version in which SSL distillation is not performed, and the original UniFormer weights pre-trained on general domain are directly used for Stage II. 
PRISM without Stage II represents an ablation version of our proposed model in which the visual features from Stage I are directly fed into survival analysis.

\begin{table}[h!]
  \centering
  \caption{Ablation study results (C-index) comparing PRISM with and without Stage II across cohorts under IECV settings.}
  \resizebox{\textwidth}{!}{%
    \begin{tabular}{lcccc}
    \toprule
    Model & GLCCM & AZCCM & RJCCM & TJCCM \\
    \midrule
          & \multicolumn{4}{c}{\textbf{Internal}} \\
    \midrule
    PRISM (Ours, w/o Stage I) & 0.635$\pm$0.120 & 0.762$\pm$0.024 & 0.601$\pm$0.033 & 0.704$\pm$0.096 \\
    PRISM (Ours, w/o Stage II) & 0.600$\pm$0.031 & 0.671$\pm$0.049 & 0.615$\pm$0.072 & {0.730}$\pm$0.079 \\
    PRISM (Ours, w/ Stage I, II) & \textbf{0.690}$\pm$0.055 & \textbf{0.772}$\pm$0.035 & \textbf{0.617}$\pm$0.033 & \textbf{0.775}$\pm$0.037 \\
    \midrule
          & \multicolumn{4}{c}{\textbf{External}} \\
    \midrule
    PRISM (Ours, w/o Stage I) & 0.702$\pm$0.078 & 0.687$\pm$0.032 & 0.528$\pm$0.049 & 0.581$\pm$0.093 \\
    PRISM (Ours, w/o Stage II) & {0.617}$\pm$0.061 & 0.679$\pm$0.047 & \textbf{0.665}$\pm$0.088 & {0.608}$\pm$0.115 \\
    PRISM (Ours, w/ Stage I, II) & \textbf{0.710}$\pm$0.010 & \textbf{0.712}$\pm$0.031 & {0.609}$\pm$0.077 & \textbf{0.701}$\pm$0.075 \\
    \bottomrule
    \end{tabular}%
  }
  \label{tab:ablation_c-index}%
\end{table}%

\begin{table}[h!]
  \centering
  \caption{Ablation study results (AUC) comparing PRISM with and without Stage II across cohorts under IECV settings.}
  \resizebox{\textwidth}{!}{%
    \begin{tabular}{lcccc}
    \toprule
    Model & GLCCM & AZCCM & RJCCM & TJCCM \\
    \midrule
          & \multicolumn{4}{c}{\textbf{Internal}} \\
    \midrule
    PRISM (Ours, w/o Stage I) & 0.634$\pm$0.123 & 0.705$\pm$0.037 & 0.583$\pm$0.045 & 0.708$\pm$0.107 \\
    PRISM (Ours, w/o Stage II) & 0.606$\pm$0.052 & 0.667$\pm$0.055 & 0.592$\pm$0.068 & {0.744}$\pm$0.081 \\
    PRISM (Ours, w/ Stage I, II) & \textbf{0.689}$\pm$0.036 & \textbf{0.714}$\pm$0.016 & \textbf{0.593}$\pm$0.030 & \textbf{0.760}$\pm$0.066 \\
    \midrule
          & \multicolumn{4}{c}{\textbf{External}} \\
    \midrule
    PRISM (Ours, w/o Stage I) & \textbf{0.698}$\pm$0.092 & 0.653$\pm$0.044 & 0.528$\pm$0.076 & 0.577$\pm$0.107 \\
    PRISM (Ours, w/o Stage II) & {0.622}$\pm$0.057 & 0.651$\pm$0.059 & \textbf{0.663}$\pm$0.092 & 0.568$\pm$0.148 \\
    PRISM (Ours, w/ Stage I, II) & {0.692}$\pm$0.019 & \textbf{0.676}$\pm$0.028 & 0.589$\pm$0.091 & \textbf{0.657}$\pm$0.130 \\
    \bottomrule
    \end{tabular}%
  }
  \label{tab:ablation_auc}%
\end{table}%

To enable PRISM modeling in the out-domain setting, we harmonized the categorical structure of EHR features and examined the effect of applying or omitting strata stratification, allowing the CoxPH survival head to adaptively learn a set of basis functions from the training data. As shown in Table~\ref{tab:out-domain_c-index} and Table~\ref{tab:out-domain_auc}, we systematically evaluated different feature selection strategies (stepwise regression and Lasso regularization) combined with strata stratification across all four cohorts under external validation settings.

\begin{table}[h!]
  \centering
  \caption{Survival analysis results of the PRISM model using prompt defined in the main table, evaluated under different feature selection strategies and strata setting. Abbreviations: Str, strata; Ste, stepwise; L, lasso. Metrics: C-index.}
  \resizebox{\textwidth}{!}{%
    \begin{tabular}{ccccccc}
    \toprule
    Str   & Ste   & L     & GLCCM C-index & AZCCM C-index & RJCCM C-index & TJCCM C-index \\
    \midrule
          &       &       & 0.708$\pm$0.016 & 0.687$\pm$0.030 & 0.602$\pm$0.077 & 0.596$\pm$0.110 \\
          &       & \checkmark     & 0.709$\pm$0.011 & 0.691$\pm$0.028 & 0.590$\pm$0.069 & 0.647$\pm$0.108 \\
          & \checkmark     &       & 0.675$\pm$0.036 & 0.678$\pm$0.008 & 0.574$\pm$0.063 & 0.658$\pm$0.078 \\
    \checkmark     &       & \checkmark     & 0.710$\pm$0.010 & 0.712$\pm$0.031 & 0.609$\pm$0.077 & 0.701$\pm$0.075 \\
    \checkmark     & \checkmark     &       & 0.676$\pm$0.035 & 0.659$\pm$0.032 & 0.591$\pm$0.058 & 0.691$\pm$0.060 \\
    \bottomrule
    \end{tabular}%
  }
  \label{tab:out-domain_c-index}%
\end{table}%

\begin{table}[h!]
  \centering
  \caption{Survival analysis results of the PRISM model using prompt defined in the main table, evaluated under different feature selection strategies and strata setting. Abbreviations: Str, strata; Ste, stepwise; L, lasso. Metrics: AUC.}
  \resizebox{\textwidth}{!}{%
    \begin{tabular}{ccccccc}
    \toprule
    Str   & Ste   & L     & GLCCM AUC & AZCCM AUC & RJCCM AUC & TJCCM AUC \\
    \midrule
          &       &       & 0.696$\pm$0.030 & 0.651$\pm$0.041 & 0.590$\pm$0.097 & 0.589$\pm$0.132 \\
          &       & \checkmark     & 0.699$\pm$0.020 & 0.654$\pm$0.028 & 0.583$\pm$0.082 & 0.615$\pm$0.119 \\
          & \checkmark     &       & 0.667$\pm$0.023 & 0.652$\pm$0.020 & 0.576$\pm$0.086 & 0.634$\pm$0.097 \\
    \checkmark     &       & \checkmark     & 0.692$\pm$0.019 & 0.676$\pm$0.028 & 0.589$\pm$0.091 & 0.657$\pm$0.130 \\
    \checkmark     & \checkmark     &       & 0.665$\pm$0.023 & 0.627$\pm$0.028 & 0.588$\pm$0.070 & 0.660$\pm$0.109 \\
    \bottomrule
    \end{tabular}%
  }
  \label{tab:out-domain_auc}%
\end{table}%

 We observed that enabling strata stratification consistently improved the survival analysis performance of PRISM under out-domain conditions. Specifically, the combination of strata stratification with Lasso regularization (Str + L) achieved the best performance in most cohorts, with C-index values of 0.710, 0.712, 0.609, and 0.701 for GLCCM, AZCCM, RJCCM, and TJCCM, respectively. This finding underscores the importance of adaptive basis function learning in the CoxPH model for handling distribution shifts across different clinical centers.

\subsection{Individualized PRISM-based risk interpretation}
\begin{figure}[h!]
    \centering
    \includegraphics[width=1\linewidth]{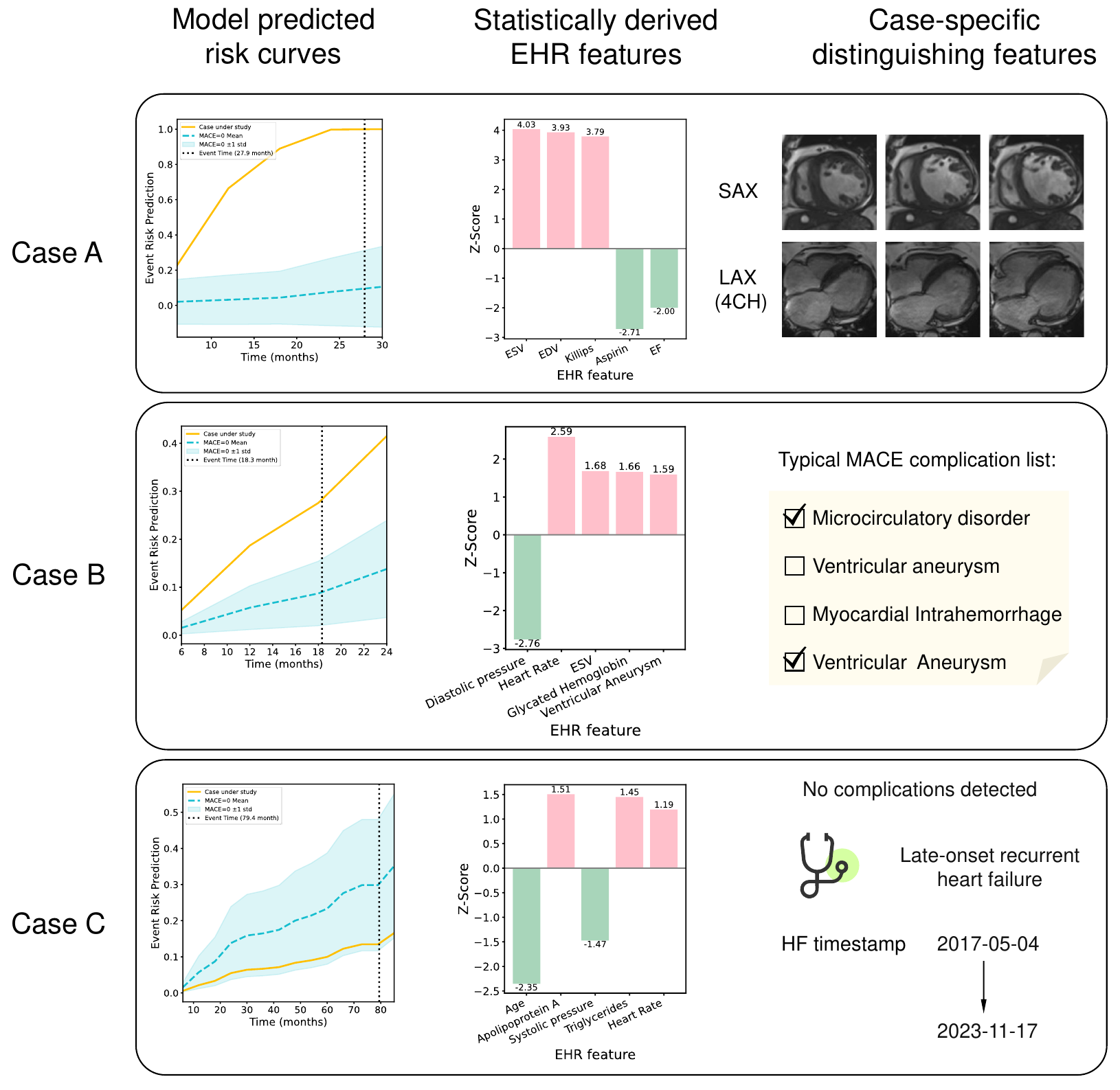}
    \caption{Case-specific risk profiling using PRISM. A teal color and an amber color are used to illustrate the temporal risk prediction trends of PRISM for the case under study and a MACE=0 sample from the same cohort, respectively. Statistically derived EHR features and case-specific distinguishing features are listed to highlight key evidence associated with the sample's survival analysis.}
    \label{fig: case_study}
\end{figure}

A case study was conducted to illustrate PRISM risk estimates for event occurrence at successive time points within the same cohort, comparing representative MACE-positive patients with the mean profile of MACE-negative patients. As Fig.~\ref{fig: case_study} shows, physician-curated EHR features supported the analysis and interpretation of these cases, elucidating their complication profiles and salient clinical characteristics.

Case A and case B demonstrate PRISM's risk predictions over approximately one year in typical high-risk cases. A more pronounced divergence in the risk curves is observed in case A, where PRISM's risk prediction reached a significantly elevated level as early as 20 months post-prognosis. This can be partly attributed to the distinct quantitative differences in ventricular motion characteristics, such as markedly elevated end-systolic volume (ESV) and end-diastolic volume (EDV), coupled with a notably reduced EF. From the 25-frame multi-view imaging series, we selected the 1st, 11th, and 21st frames for analysis. Both SAX and LAX views reveal significantly enlarged ventricular volumes with minimal chamber variation throughout the cardiac cycle, corroborating the imaging-based evidence of severely impaired EF.

In contrast, among the listed complications strongly associated with MA-\allowbreak CE, case B exhibited two conditions, including microcirculatory disorder and ventricular aneurysm. Compared to MACE=0 samples, its differential EHR feature ranking showed less pronounced divergence than in case A, as reflected in the smaller Z-score deviation from zero. This observation may partly explain why PRISM adopted a more conservative approach in predicting MACE for case B.

Case C represents the most misleading scenario in our study, specifically selected to demonstrate the model's potential for prediction errors. This patient experienced a recurrent heart failure leading to MACE after 78 months of follow-up, despite displaying no high-risk indicators during initial assessment. Notably, baseline characteristics including age and apolipoprotein A levels were more favorable than those observed in typical cases.
This discordance between model predictions and actual outcomes emphasizes the critical importance of extended clinical surveillance, as early prognostic assessments may fail to identify delayed adverse events.

\section{Discussion}
A key finding of our study is the striking concordance between the spatiotemporal risk signatures identified by PRISM and established coronary artery territories. It is crucial to emphasize that this alignment emerged de novo from the model's unsupervised learning process, without any prior anatomical or physiological knowledge of vascular distribution being encoded. This convergence is particularly significant given that myocardial infarction, a prototypical coronary artery disease, underlies the pathology of our study cohort. Specifically, the inferior hypersensitivity signature likely reflects residual dysfunction within the territory supplied by the right coronary artery, while the dynamic patterns of lateral dyssynchrony and anterior diastolic elevated focus correspond to regions perfused by the left anterior descending and circumflex arteries, respectively. This data-driven recapitulation of fundamental clinical pathophysiology serves as a tool capable of generating new hypotheses about how specific patterns of regional myocardial dysfunction contribute to the overall prognosis following an acute ischemic event, which elevates the clinical potential of our approach beyond simple risk stratification.

Integrating the cross-cohort SHAP value analyses derived from our PRISM model, we identified that among routinely obtained clinical assessments in standard hospital settings, hypertension and diabetes emerge as prominent potential biomarkers with high priority for predicting MACE. Notably, these comorbidities consistently ranked within the top-10 salient features across three of our independent centers. This finding underscores the persistent and critical role of traditional EHR variables alongside the advanced, unsupervised, image-based feature extraction facilitated by PRISM in survival analyses. 

This finding has been corroborated in medical studies focused on EHR-based data mining. For instance, \cite{Chopannejad2022,Zhai2025} identified these three factors as having odds ratios substantially deviating from 1 among a wide range of EHR-derived features. Some studies\cite{Alison2019Cardiovascular, Tao2023} directly targeted patient populations with the corresponding conditions to investigate cardiovascular risk. However, such studies are predominantly based on aggregated clinical statistics, and few have initiated the investigation from high-dimensional medical imaging before incorporating multimodal clinical features.
On the other hand, while EHR data contribute substantially to predictive performance, their center-specific nature raises important concerns regarding generalizability. Certain EHR-derived predictors previously established as pivotal for MACE prognostication, such as door-to-balloon (D2B) time~\cite{Huang2023InHospitalMACE}, which may exhibit limited external validity due to inter-institutional variability and data scarcity. This heterogeneity accentuates the imperative for robust image-derived biomarkers and the refinement of methodologies for their extraction and integration, as these features potentially offer more scalable and universally applicable risk indicators. Ultimately, our findings advocate for a hybrid modeling paradigm that synergizes comprehensive EHR information with advanced imaging analytics to achieve optimal, generalizable cardiovascular risk assessment.

In addition to the population-level findings, the case study indicated that patients who experienced MACE frequently presented with complications such as microcirculatory disorder and ventricular thrombosis. Although these conditions may represent intermediate stages in the development of MACE, they do not constitute decisive determinants of the event itself.

\section{Conclusion}
PRISM demonstrates superior MACE survival prediction performance across four independent clinical cohorts, consistently outperforming classical and SOTA baselines under both internal and external validation. The framework's unsupervised learning paradigm identifies three spatiotemporal imaging signatures, including lateral wall dyssynchrony, inferior wall hypersensitivity, and anterior diastolic elevated focus, that align with established coronary artery territories, validating its capacity to recapitulate pathophysiological principles without explicit anatomical supervision. Combined with prompt-guided attribution revealing hypertension, diabetes, and smoking as dominant EHR risk factors, PRISM establishes a practical, annotation-free approach for cardiovascular risk stratification that bridges high-dimensional imaging analytics with clinical assessments.

\section*{Declarations}
\subsection*{Ethics approval}
The study protocol was approved by the Institutional Review Board of Shanghai Renji Hospital (approval No. RJRE[2018]093). Collaborating centers, including Beijing Anzhen Hospital, Nanjing Drum Tower Hospital, and Wuhan Tongji Hospital, participated under this approved protocol. To protect patient privacy, informed consent was waived by the IRB above, and all data were anonymized prior to their use in model training, testing, and reader studies. This study was conducted in accordance with the Declaration of Helsinki and relevant local regulations.

\subsection*{Code availability}
The source code used to generate the results presented in this study is available on GitHub at https://github.com/Hoyant-Su/PRISM. The repository includes scripts for data pre-processing, the three-stage training process, and evaluation of the model.

\subsection*{Competing interests}
The authors declare no competing interests.

\subsection*{Author contribution}
H.S. designed  the model and performed experiments. H.S. and S.R. developed and provided survival analysis code. H.S., S.R., and Y.G. processed the experimental data and documented control experiment details. J.X., X.C., T.Y., and L.W. provided cine imaging data, medical diagnoses, and electronic health records. S.Z., L.W., and X.W. supervised the research direction. H.S. drafted the manuscript. J.X., L.W., and X.W. contributed to manuscript editing and revision.

\subsection*{Corresponding author}
Correspondence to Lian-Ming Wu and Xiaosong Wang.

\subsection*{Funding}
This research received funding from 
the National Program for Support of Top-notch Young Professionals from the Organization Department of the Central Committee of the Communist Party of China, 
the National Natural Science Foundation of China (No. 82171884, No. 82471931), 
the Shanghai Municipal Commission of Science and Technology Medical Innovation Research Special Project (23Y11906900), 
and the Shanghai Municipal Health Commission Outstanding Youth Grant.

\subsection*{Acknowledgements}
This work received support from the Shanghai Innovation Institute and the Shanghai Artificial Intelligence Laboratory (No. PJ-PRJ24AI4S001).

%% If you have bib database file and want bibtex to generate the
%% bibitems, please use
%%
 \bibliographystyle{elsarticle-num} 
 \bibliography{sn-bibliography}

%% else use the following coding to input the bibitems directly in the
%% TeX file.

%% Refer following link for more details about bibliography and citations.
%% https://en.wikibooks.org/wiki/LaTeX/Bibliography_Management

% \begin{thebibliography}{00}

% %% For numbered reference style
% %% \bibitem{label}
% %% Text of bibliographic item

% \bibitem{lamport94}
%   Leslie Lamport,
%   \textit{\LaTeX: a document preparation system},
%   Addison Wesley, Massachusetts,
%   2nd edition,
%   1994.

% \end{thebibliography}
\end{document}